\documentclass[conference]{IEEEtran}
\IEEEoverridecommandlockouts
\usepackage{cite}
\usepackage{amsmath,amssymb,amsfonts}
\usepackage{algorithmic}
\usepackage{graphicx}
\usepackage{textcomp}
\usepackage{xcolor}
\def\BibTeX{{\rm B\kern-.05em{\sc i\kern-.025em b}\kern-.08em
    T\kern-.1667em\lower.7ex\hbox{E}\kern-.125emX}}
    
\usepackage{booktabs} 
\usepackage{microtype}
\usepackage{graphicx}
\usepackage{subfigure}
\usepackage{booktabs} 
\usepackage{makecell}
\usepackage{hyperref}
\usepackage{pdflscape}
\usepackage{mathrsfs}
\usepackage{paralist}
\usepackage{multirow}
\usepackage{enumitem}
\usepackage{amsmath}
\usepackage{bbold}
\usepackage{mathtools}
\usepackage[normalem]{ulem}
\usepackage{amsthm}

\usepackage{cuted}
\usepackage{flushend}

\usepackage{url}
\usepackage{makecell}

\newcommand{\bit}{\begin{compactitem}}
\newcommand{\eit}{\end{compactitem}}
\newcommand{\ben}{\begin{compactenum}}
\newcommand{\een}{\end{compactenum}}
\newcommand{\mytag}[1]{{\bf#1}}

\newcommand{\hide}[1]{}

\newcommand{\model}{SCEHR\ }
\newcommand{\modelEOL}{SCEHR}
\newcommand{\modela}{{$\mathcal{L}_{\text{CBCE}} + \lambda \mathcal{L}_{\text{SCR}}$} }
\newcommand{\modelb}{{$\mathcal{L}_{\text{CSCE}} + \lambda \mathcal{L}_{\text{SCR}}$} }
\newcommand{\code}{\url{https://github.com/calvin-zcx/SCEHR}}

\DeclareMathOperator*{\argmin}{arg\,min}

\usepackage[ruled,vlined]{algorithm2e}

\theoremstyle{definition}

\theoremstyle{remark}

\begin{document}

\title{
SCEHR: Supervised Contrastive Learning for Clinical Risk Prediction using Electronic Health Records
}

\author{\IEEEauthorblockN{Chengxi Zang} 
\IEEEauthorblockA{\textit{Population Health Sciences} \\
\textit{Weill Cornell Medicine}\\
New York, NY, USA \\
chz4001@med.cornell.edu}
\and
\IEEEauthorblockN{Fei Wang} 
\IEEEauthorblockA{\textit{Population Health Sciences} \\
\textit{Weill Cornell Medicine}\\
New York, NY, USA \\
few2001@med.cornell.edu}
}
\maketitle

\begin{abstract}
Contrastive learning has demonstrated promising performance in image and text domains either in a self-supervised or a supervised manner. In this work, we extend the supervised contrastive learning framework to clinical risk prediction problems based on longitudinal electronic health records (EHR). We propose a general supervised contrastive loss $\mathcal{L}_{\text{Contrastive Cross Entropy} }  + \lambda \mathcal{L}_{\text{Supervised Contrastive Regularizer}}$ for learning both binary classification (e.g. in-hospital mortality prediction)
and multi-label classification (e.g. phenotyping) in a unified framework. 
Our supervised contrastive loss practices the key idea of contrastive learning, namely, pulling similar samples closer and pushing dissimilar ones apart from each other, simultaneously by its two components:  $\mathcal{L}_{\text{Contrastive Cross Entropy} }$ tries to contrast  samples with learned anchors which represent positive and negative clusters, and $\mathcal{L}_{\text{Supervised Contrastive Regularizer}}$ tries to contrast samples with each other according to their supervised labels.
We propose two versions of the above supervised contrastive loss and our experiments on real-world EHR data demonstrate  that our proposed loss functions show benefits in improving the performance of strong  baselines and even state-of-the-art models on benchmarking tasks for clinical risk predictions. Our loss functions work well with extremely imbalanced data which are common for clinical risk prediction problems. Our loss functions can be easily used to replace (binary or multi-label) cross-entropy loss adopted in existing clinical predictive models. The Pytorch code is released at \code .
\end{abstract}

\begin{IEEEkeywords}
Supervised contrastive learning; Supervised contrastive loss; Contrastive cross entropy; Supervised contrastive regularizer; Clinical risk predictions; Electronic Health Records; Clinical time series; In-hospital mortality prediction; Phenotyping; Multi-label classification
\end{IEEEkeywords}

\section{Introduction}
\label{sec:intro}
With the accumulation and better availability of electronic health records (EHR) \cite{miotto2018deep, solares2020deep}, health analytics becomes one of the most important frontiers for data mining and artificial intelligence \cite{wang2019ai}. Public EHR databases \cite{johnson2016mimic} and benchmark suite \cite{harutyunyan2019multitask} provide great resource to develop advanced data mining and machine learning algorithms for critical clinical risk prediction problems including in-hospital mortality prediction, disease phenotyping, hospital readmission, etc. \cite{harutyunyan2019multitask, rajkomar2018scalable}. These problems can be formulated as a binary or multi-label classification problem using longitudinal EHR event sequence (by concatenating visits of individual patients over time) and solved by minimizing its corresponding classification loss [e.g. (multi-label or binary) cross-entropy loss]\cite{harutyunyan2019multitask, rajkomar2018scalable, wanyan2021contrastive}. Although great endeavors have been devoted to developing complex deep learning models for these clinical risk prediction problems \cite{harutyunyan2019multitask, gao2020stagenet, li2020behrt, kodialam2020deep, chen2020unite, ma2020adacare, choi2016retain, song2018attend, ma2020concare, luo2020hitanet, bellamy2020evaluating}, 
limited progress has been made over past years on these tasks regarding their performance \cite{bellamy2020evaluating}. In contrast with the majority of current research in designing  more advanced predictive models, in this paper, we show that replacing widely adopted  cross entropy loss by  supervised contrastive loss is a promising way to improve the performance of existing models for clinical risk prediction  based on longitudinal EHR data.

\begin{figure*}[!tb]
\centering
\includegraphics[width=0.73\textwidth]{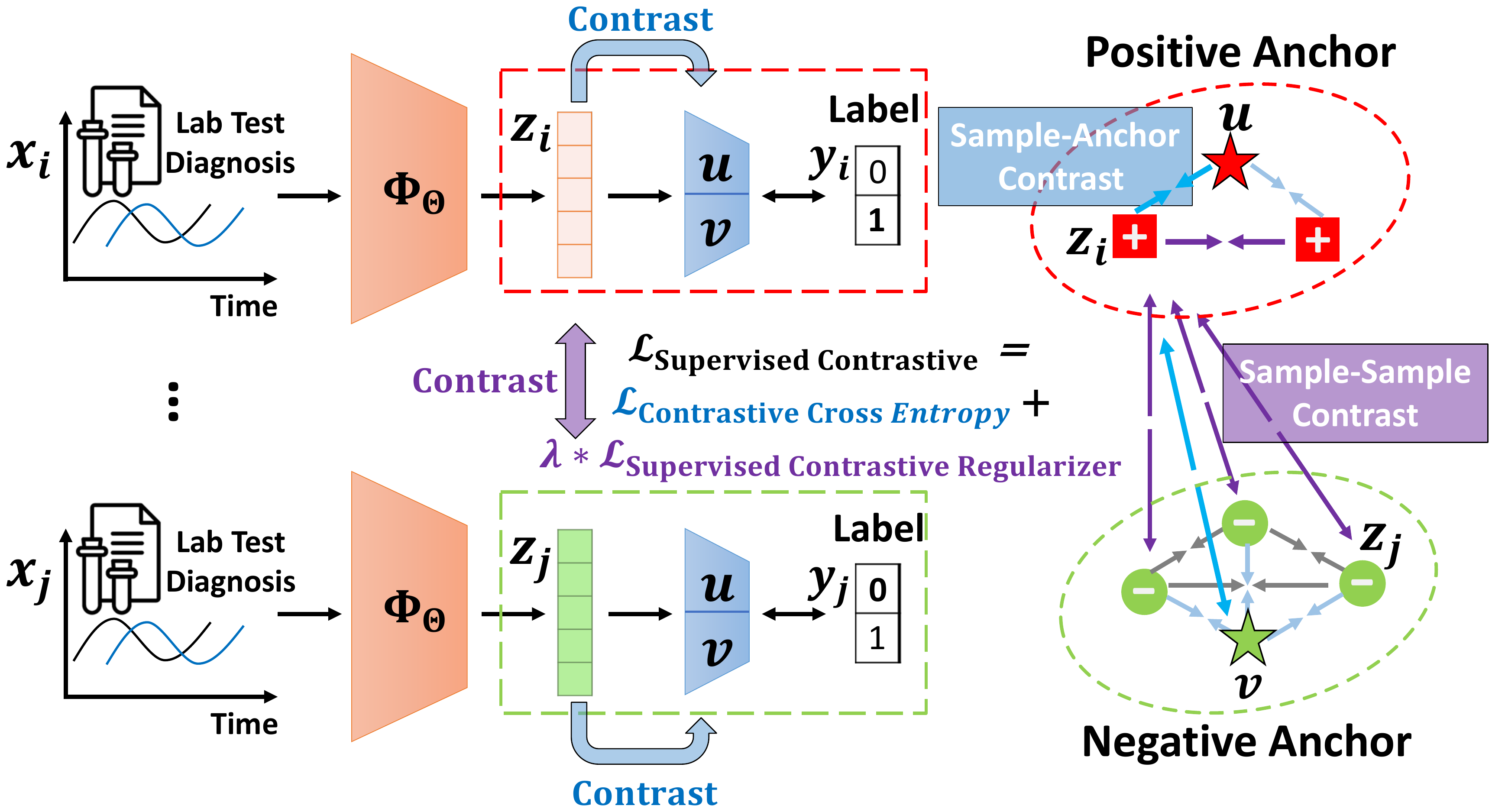}
\caption{ 
An illustration of our \modelEOL. We propose a general supervised contrastive learning loss   $\mathcal{L}_{\text{Contrastive Cross Entropy} }  + \lambda \mathcal{L}_{\text{Supervised Contrastive Regularizer}}$  for  clinical risk prediction problems using longitudinal electronic health records. The overall goal is to improve the performance of binary classification (e.g. in-hospital mortality prediction) and multi-label classification (e.g. phenotyping) by pulling ($\rightarrow \leftarrow $) similar samples closer and pushing ($ \leftarrow \rightarrow$) dissimilar samples apart from each other. $\mathcal{L}_{\text{Contrastive Cross Entropy} }$ tries to contrast  sample representations with learned positive and negative anchors, and $\mathcal{L}_{\text{Supervised Contrastive Regularizer}}$ tries to contrast sample representations with others in a mini-batch according to their labels. For brevity, we only highlight the contrastive pulling and pushing forces associated with sample $i$ in a mini-batch consisting of two positive samples and three negative samples.
\label{fig:model}}
\end{figure*}

Recently, contrastive learning \cite{le2020contrastive}, which aims at learning data instance representations by bringing similar instances closer and push dissimilar instances further away from each other, has shown promising results in image classifications \cite{chen2020simple, caron2020unsupervised}, medical image understanding \cite{zhang2020contrastive}, and so on \cite{liu2020self}. These methods mainly follow a self-supervised strategy \cite{liu2020self, jaiswal2021survey}, which build augmented data with pseudo-labels to deal with the issue of lacking sufficient supervised information. 
The latest research finds that supervised information can provide additional benefits for contrastive learning in both computer vision \cite{khosla2020supervised} and natural language processing tasks \cite{gunel2020supervised}. We argue that the general idea of contrastive learning should also be helpful for clinical risk prediction tasks. However, application of contrastive learning in clinical risk prediction scenarios is challenging because: 1) the patient data (such as EHRs) for clinical risk prediction are usually 
more complex than images or texts in that the clinical events involved are of mixed types, high-dimensional, sparse and noisy; 2) 
it is challenging to augment EHR with computational methods because of the intrinsic complexity of disease mechanisms; 3) 
 predicted clinical outcomes could also be heterogeneous. 
Therefore, if contrastive learning strategies can be beneficial to clinical risk prediction problems is still an open question. 
 
In this paper, we propose \mytag{\modelEOL}, a \mytag{\underline{S}}upervised \mytag{\underline{C}}ontrastive learning framework for clinical risk predictions using longitudinal \mytag{\underline{E}}lectronic \mytag{\underline{H}}ealth \mytag{\underline{R}}ecord data. We illustrate the idea of  \model in Figure~\ref{fig:model}. The key component of \model is \mytag{a general supervised contrastive loss} $\mathcal{L}_{\text{Supervised Contrastive} } = \mathcal{L}_{\text{Contrastive Cross Entropy} }  +  \\ \lambda \mathcal{L}_{\text{Supervised Contrastive Regularizer}} $ for solving binary classification (e.g. in-hospital mortality prediction)
and multi-label classification (e.g. phenotyping) in a unified framework. We propose two versions (Eq.~\ref{eq:multi-cbce} and Eq.~\ref{eq:multi-csce}) of the above supervised contrastive loss to implement the key idea of contrastive learning, i.e., pulling similar samples closer and pushing dissimilar ones apart from each other, which can be achieved by minimizing the two components of our $\mathcal{L}_{\text{Supervised Contrastive} }$. Specifically, for an arbitrary  neural encoder that maps  clinical time series into embedding representations, the $\mathcal{L}_{\text{Contrastive Cross Entropy} }$  learns a positive  anchor  and a negative anchor (for each class) respectively and tries to contrast the distance between targeted samples and the learned positive anchor versus the distance between the targeted samples and the learned negative anchor, guided by the supervised labels (e.g. positive/dead for in-hospital mortality prediction, or existence of some medical concepts for phenotyping classification). The $\mathcal{L}_{\text{Supervised Contrastive Regularizer}}$ tries to contrast  every pair of samples with the same labels versus every pair of samples with different labels in a mini-batch. 
By leveraging supervised information, \model doesn't need data augmentation and pseudo-labels. In addition, we also demonstrate the relationship between $\mathcal{L}_{\text{Supervised Contrastive} }$ and the triplet loss \cite{chechik2010large}. 

We validate \model together with two versions of our proposed supervised contrastive losses on benchmarking clinical risk prediction tasks, including in-hospital mortality prediction and phenotyping \cite{harutyunyan2019multitask}, on a big real-world EHR database (MIMIC-III) \cite{johnson2016mimic}. We find that both versions of our proposed loss functions can improve strong baseline models and state-of-the-art models. We further investigate our modeling performance when the level of data imbalance changes. We find that our proposed loss functions work much better than binary cross entropy loss under extreme imbalance situation (say, positive ratio $\le 1 \%$), which is common in prediction problems with rare clinical outcomes. We further visualize our learned embeddings to interpret the effects of our proposed supervised contrastive losses.
It is worthwhile to highlight our contributions as follows:
  \bit
 \item {\bf Novelty.} We propose a general supervised contrastive loss $\mathcal{L}_{\text{Contrastive Cross Entropy} }  +  \lambda \mathcal{L}_{\text{Supervised Contrastive Regularizer}} \ $ and its two instances for solving supervised binary classification and multi-label classification in a unified framework. \model is one of the first applying \textit{supervised contrastive learning} to clinical risk predictions with longitudinal EHR data.
 
\item {\bf Effectiveness.} \model can improve both strong baseline models and the state-of-the-art models for clinical risk prediction tasks, including in-hospital mortality prediction and phenotyping. \model does well with extreme data imbalance situation. 

\item {\bf Flexibility.} Our proposed supervised contrastive loss functions can be easily used to replace (multi-label or binary) cross entropy loss based on existing clinical predictive models. Our PyTorch code is open-sourced  at \code.
\eit

The outline of this paper is: 
survey (Sec.~\ref{sec:related}), problem definition (Sec.~\ref{sec:problem}), proposed method \model (Sec.~\ref{sec:model}), experiments (Sec.~\ref{sec:exp}), and conclusions (Sec.~\ref{sec:conclusion}).  

\section{Related Work}
\label{sec:related}

\mytag{Deep predictive models using EHR data.}
Applying deep models for clinical risk prediction problems (e.g. in-hospital mortality prediction, phenotyping, decompensation, length-of-stay prediction, readmissions, etc.) based on longitudinal electronic health record (EHR) data \cite{miotto2018deep, rajkomar2018scalable, solares2020deep} show great potentials in improving health care. These tasks are usually formulated as binary or multi-label classification problems by optimizing multi-label or binary cross-entropy loss. Most of research endeavors have been devoted to developing more advanced deep models or trying to incorporate more data to capture the complexity of diseases and the EHR data, including but not limited to RNNs \cite{harutyunyan2019multitask, gao2020stagenet}, transformers \cite{li2020behrt}, reverse distillation \cite{kodialam2020deep}, variational inference \cite{chen2020unite}, deep feature selection \cite{ma2020adacare}, attentions \cite{choi2016retain, song2018attend, ma2020concare, luo2020hitanet}, an so on. However, despite the fast pace of modeling innovations, much slower progress has been made over past years on these tasks concerning their performance \cite{bellamy2020evaluating}.
Instead of designing more complex deep predictive models, here we explore another direction: trying to innovate the default (binary or multi-label) cross entropy loss widely used in existing clinical predictive models. We focus on state-of-the-art models \cite{harutyunyan2019multitask, ma2020concare, bellamy2020evaluating} which were benchmarked on public   MIMIC-III data \cite{johnson2016mimic} considering limitations of using private EHR data.



\noindent \mytag{Contrastive Learning.}
Contrastive learning \cite{le2020contrastive, liu2020self}, aiming at learning good representations by bringing similar samples closer and push dissimilar samples away from each other through constructing contrastive loss functions, has shown promising results in image classifications \cite{chen2020simple, caron2020unsupervised}, medical image understanding \cite{zhang2020contrastive}, videos \cite{han2020self}, etc. The idea of "contrastive" loss functions can date back to metric learning \cite{weinberger2009distance}, triplet loss \cite{chechik2010large},  Siamese neural networks \cite{chicco2021siamese}, and the negative sampling loss of word2vec \cite{mikolov2013distributed}. 
The majority of contrastive learning literature adopted self-supervised techniques \cite{liu2020self, jaiswal2021survey, kalantidis2020hard, li2020contrastive} by building augmented data with pseudo-labels. Recently, by explicitly using supervised labels, supervised contrastive learning has shown better performance for image classification \cite{khosla2020supervised} and NLP tasks \cite{gunel2020supervised}. To our best knowledge, only one paper \cite{wanyan2021contrastive} tried the contrastive idea for binary classification with EHR data, which adopted the negative sampling loss of word2vec \cite{mikolov2013distributed} by negatively sampling on built heterogeneous information networks \cite{choi2017gram}. 
Different from all the above research, we propose a general supervised contrastive loss (together with its two versions) for solving binary classification and multi-label classification in a unified framework using longitudinal EHR data.

\section{Problem Definition}
\label{sec:problem} 
In this section, we define our focused clinical risk prediction problems with longitudinal electronic health records (EHR) data.
Let $x_{i}  \in \mathbb{R}^{T_i \times D}$ represent one patient's clinical time series data, which consist of $D$-dimensional clinical concepts  (e.g. individual measurements during his/her stay in ICU) over time  $T_i$. Specifically, $x_{i, t, d} \in \mathbb{R}$ represents the $d^{th} \in \{1, 2, ..., D\}$ clinical concept (e.g. diastolic blood pressure) measured at timestamp $t \in \{1, 2, ..., T_i\}$ for patient $i$. 
In total, there are $N$  patients denoted as $X = \{x_1, x_2, ..., x_N\}$ and $T_i$ ($i \in \{1, 2, ..., N\}$) usually varies for different patients according to their length of  stay, say, in ICU.  Additional static features, e.g. demographic features, are denoted as $S \in \mathbb{R}^{N \times D_S} $ and $s_i \in \mathbb{R}^{1 \times D_S} $ represents patient $i$'s features. 
For simplicity, we use $X = (X, S)$ to represent all the clinical time series and additional static features (if exist) for modeling.
We use $Y \in \{0,1\} ^ {N \times D_Y} $ to denote the targeted clinical outcomes, e.g. in-hospital mortality events,  the existence of phenotype conditions, etc., which will occur beyond the observational window $T_i$ ($i \in \{1, 2, ..., N\}$) for each patient, and $D_Y \in \mathbb{N}^+$. 

Our primary goal is to learn a predictive model $\mathcal{F}_\Theta : X \to Y $, which predicts the probability of the occurrence of clinical outcomes denoted as $\hat{Y}$. The $\Theta$ are learnable modeling parameters. Regarding the value of $D_Y$, the above problem formulation encompasses two special cases:
\bit
\item \mytag{Binary classification problem $(D_Y=1)$}, namely,  $\mathcal{F}_\Theta : X \to Y $ where $Y \in \{0,1\} ^ {N \times 1}$. Tasks including in-hospital mortality prediction, physiologic decompensation, etc., belong to this category.
\item \mytag{Multi-label classification problem $(D_Y > 1)$}, namely,  $\mathcal{F}_\Theta : X \to Y $ where $Y \in \{0,1\} ^ {N \times D_S}$, which  can be formulated as solving multiple binary classifications simultaneously. The  phenotype classification (phenotyping) task belongs to this category.
\eit
We will detail the above tasks in the experiment sections. We learn the parameters $\Theta$ of $\mathcal{F}_\Theta$ by minimizing the loss function:
\begin{equation} 
\argmin_{\Theta}  \ \mathcal{L} (\mathcal{F}_\Theta(X), Y)
\end{equation}
given supervised information $Y$, and $\hat{Y} = \mathcal{F}_\Theta(X)$ are the predicted outcomes.

In contrast with the majority of existing efforts in designing  $\mathcal{F}_\Theta$, in this paper, we show that the supervised contrastive learning loss  $\mathcal{L}_{\text{supervised contrastive}}$  proposed as follows is also an effective way to improve the performance of clinical predictive models.

\section{Supervised Contrastive Learning Framework for EHR}
\label{sec:model}
In this section, we introduce our \underline{S}upervised \underline{C}ontrastive Learning for \underline{EHR}  ({\modelEOL}) model in detail. We show the outline of our \model in Figure~\ref{fig:model} as a roadmap for this section and we summarize the overall learning process of our \model  in Algorithm~\ref{alg:scehr}.

\subsection{General Supervised Contrastive Loss}
\label{sec:model:general}
 Let $\Phi_{\Theta}$ be any learnable  neural encoder for clinical time series $X$, which maps $X$ into its embedding representation $Z$ by $Z = \Phi_{\Theta}(X)$. We further define a linear mapping $f$ and a non-linear squeeze function $\sigma$ (e.g. sigmoid or softmax functions)  which maps the learned representations to the predicted probability by $\hat{Y}=\sigma \circ f(Z)$. 
We propose the following general form of \textit{Supervised Contrastive Loss} for binary or multi-label classification problems:
\begin{equation}
\label{eq:scl-main}
\begin{aligned}
\mathcal{L}_{\text{Supervised Contrastive} }  = \mathcal{L}_{\text{Contrastive Cross Entropy} }  + \\
\lambda \mathcal{L}_{\text{Supervised Contrastive Regularizer}} 
\end{aligned}
\end{equation}
Our $\mathcal{L}_{\text{Supervised Contrastive} }(\hat{Y}, Z, Y)$ loss consists of two parts: a (supervised) contrastive cross entropy loss $\mathcal{L}_{\text{contrastive cross entropy} }  $ which is a function of predicted labels $\hat{Y}$ against its ground truth labels $Y$; and a supervised contrastive regularizer $\mathcal{L}_{\text{supervised contrastive regularizer}} $ which regularizes the learned embedding representation $Z$ by the supervised information $Y$. 
The regularizer is scaled by a non-negative hyper-parameter $\lambda$. We will detail several choices of the above losses for both binary classification and multi-label classification as follows.

\subsection{Contrastive Cross Entropy for Binary Classification}
\label{sec:model:cce}

Let $x \in X$,  $z \in Z$,  $y \in Y$, and $\hat{y} \in \hat{Y}$ represent clinical time series of one patient, its embedding representation, its ground-truth clinical outcomes, and its predicted outcomes respectively. We use $u, v$ to represent the learned anchors of positive or negative clusters respectively, which are modeled as the row vectors of the weight matrix of a linear mapping $f$. 

The \mytag{Binary Cross Entropy (BCE)} loss is widely used for clinical risk classification when there are two outcomes coded as 1 or 0, say mortality for positive cases and non-mortality for negative cases. The equation for BCE loss, denoted as $\mathcal{L}_{\text{BCE}}$, is:
\begin{equation}
\small
\begin{aligned}
&\mathcal{L}_{\text{BCE}} \\
&= - \frac{1}{N}\sum_{i=1}^{N} y_i \log \hat{y_i} + (1-y_i) \log (1 - \hat{y_i})\\
&= - \frac{1}{N}\sum_{i=1}^{N} y_i \log \sigma(u^T z_i) + (1-y_i) \log (1 - \sigma(u^T z_i))\\
&= -\frac{1}{N} \sum_{i=1}^{N} y_i \log \sigma(u^T z_i) + (1-y_i) \log \sigma(-u^T z_i) \\
&= - \frac{1}{N} \sum_{i=1}^{N} y_i \log \sigma( \text{sim} (u, z_i)) + (1-y_i) \log \sigma(\text{sim} (-u, z_i))
\end{aligned}
\end{equation}
where $\sigma(x) = \frac{1}{1+e^{-x}} \in (0, 1)$ is the Sigmoid function and $1 - \sigma(x) = \frac{1}{1+e^{x}} =\sigma(-x)$. If we define a distance measure $\text{sim}(u, z_i)=u^T z_i$ as the dot product of two data samples, intuitively, minimizing the BCE loss tries to make positive samples $z_i$ ($y_i=1$) close to the anchor $u$. Similarly, for negative samples $z_i$ ( $y_i=0$), the BCE loss makes $z_i$ close to $-u$. 

Here we propose \mytag{Contrastive Binary Cross Entropy (CBCE)} loss, denoted as $\mathcal{L}_{\text{CBCE}}$, as follows:
\begin{equation}
\label{eq:cbce}
\begin{aligned}
&\mathcal{L}_{\text{CBCE}} \\ 
&= - \frac{1}{N} \sum_{i=1}^{N} y_i \log \sigma(u^T z_i)\sigma(-v^T z_i) + (1-y_i) \log \sigma(v^T z_i)\sigma(-u^T z_i) \\
& = - \frac{1}{N}\sum_{i=1}^{N} \big \{ y_i \log \sigma(\text{sim} (u, z_i))\sigma(\text{sim} (-v, z_i)) \\
& \quad \quad  \quad \quad   +  (1-y_i) \log \sigma(\text{sim} (v, z_i))\sigma(\text{sim} (-u, z_i)) \big \}
\end{aligned}
\end{equation}
which is the first version of our $\mathcal{L}_{\text{contrastive Cross Entropy} } $ term. The above $\mathcal{L}_{\text{CBCE}}$ loss explicitly learns positive anchor $u$ and negative anchor $v$ separately. Minimizing the CBCE loss makes positive sample $z_i$ (when $y_i=1$) closer to positive anchor $u$ than to the negative anchor $v$ by pulling $z_i$  closer to $u$ and at the same time pushing $z_i$ away from $v$. Similarly, for a negative sample $z_i$ (when $y_i=0$), minimizing the loss makes $z_i$ closer to negative anchor $v$ than to the positive anchor $u$ by pulling $z_i$  closer to $v$ and at the same time pushing $z_i$ away from $u$. Intuitively, two learned anchors $u$ and $v$ represent positive cluster and negative cluster respectively, and the location of each sample representation $z$ is determined by contrasting the force $\text{sim} (u, z)$ with the force $\text{sim} (u, z)$ in a product form. We show the math of these contrastive forces in the following subsection. In all,   Equation~\ref{eq:cbce} contrasts each sample with positive and negative anchors in a \textit{product form}.


 Following the similar idea of $\mathcal{L}_{\text{CBCE}}$, we can also view a two-dimensional softmax cross entropy as our second instance of the contrastive cross entropy loss $\mathcal{L}_{\text{contrastive Cross Entropy} } $. We denote  \mytag{Contrastive Softmax Cross Entropy (CSCE)} as $\mathcal{L}_{\text{CSCE}}$, which is defined by the following equation:
\begin{equation}
\label{eq:csce}
\begin{aligned}
&\mathcal{L}_{\text{CSCE}} \\ 
&= - \frac{1}{N}\sum_{i=1}^{N} \big \{y_i \log \frac{\exp{(u^T z_i)}}{\exp{(u^T z_i)} + \exp{(v^T z_i)}} \\
&\quad \quad  \quad  \quad  + (1-y_i) \log \frac{\exp{(v^T z_i)}}{\exp{(u^T z_i)} + \exp{(v^T z_i)}} \big \}\\
&= - \frac{1}{N} \sum_{i=1}^{N} \big \{ y_i \log \frac{\exp{(\text{sim} (u, z_i))}}{\exp{(\text{sim} (u, z_i))} + \exp{(\text{sim} (v, z_i))}}  \\
&\quad \quad  \quad \quad  + (1-y_i) \log \frac{\exp{(\text{sim} (v, z_i))}}{\exp{(\text{sim} (u, z_i))} + \exp{(\text{sim} (v, z_i))}} \big \} \\
\end{aligned}
\end{equation}
Equation~\ref{eq:csce} contrasts each sample with positive and negative anchors in a \textit{ratio form}, which is a two-dimensional softmax function followed by a negative likelihood loss. Taking one positive sample $z_i$ (when $y_i=1$) as an example, minimizing the above loss tries to pull $z_i$ closer to the positive anchor $u$ than to the negative anchor $v$ by pulling $z_i$ to $u$ and at the same time push $z_i$ away from $v$.

\subsection{Supervised Contrastive Regularizer}
\label{sec:model:scr}
 Compared with the  $\mathcal{L}_{\text{Contrastive Cross Entropy}}$ which compares each sample's distance to the learned positive anchor with its distance to the learned negative anchor, the  $\mathcal{L}_{\text{Supervised Contrastive Regularizer}}$ tries to explore pair-wise relationships between data samples in a mini-batch. Specifically, the $\mathcal{L}_{\text{Supervised Contrastive Regularizer}}$ tries to pull the data pairs with the same labels closer and push data pairs with different labels away from each other.
Based on the supervised contrastive loss proposed in \cite {khosla2020supervised}, we propose a simplified \mytag{Supervised Contrastive} loss as the \mytag{Regularizer (SCR)},
which is defined by the following equation:
\begin{equation}
\label{eq:scr}
\begin{aligned}
&\mathcal{L}_{\text{SCR}} (Z, Y)=\\
&\frac{-1}{N}\sum_{i=1}^{N}\frac{1}{N_{z_{i}}-1} \sum_{j=1}^{N} \mathbf{1}_{i \neq j} \mathbf{1}_{y_{i}=y_{j}} \log \frac{\exp (\text{sim}(z_i, z_j) / \tau)}{\sum_{k=1}^{N} \mathbf{1}_{i \neq k} \exp (\text{sim}(z_i, z_k) / \tau)}
\end{aligned}
\end{equation}
where $N$ is the number of samples in a mini-batch, $N_{z_i}$ is the number of samples sharing the same label as data $z_i$, $\text{sim}(z_i, z_j) = \frac{z_i^T z_j}{||z_i||||z_j||}$, and $\tau$ is the positive temperature hyper-parameter. Here we do not adopt self-supervised data augmentation strategy \cite {khosla2020supervised, chen2020simple} and we only use  existing supervised information $Y$. As a result, for each data sample $z_i$, we consider its distance to other $N-1$ samples and contrast these pair-wise distances according to if two samples share the same label as ratio form as detailed in the Equation~\ref{eq:scr}.

\subsection{Relationship with Triplet Loss}
\label{sec:model:triplet}
All the above contrastive losses $\mathcal{L}_{\text{CBCE}}$,  $\mathcal{L}_{\text{CSCE}}$ and $\mathcal{L}_{\text{SCR}}$ can be approximated by a triplet loss. As for the $\mathcal{L}_{\text{CBCE}}$, the (product form) contrastive term $\log [\sigma(u^T z)\sigma(-v^T z)] $ between sample representation $z$ and two anchors $u, v$ can be approximated as:
\begin{equation}
\begin{aligned}
 \argmin_{\Theta} & - \log  \{\sigma(u^T z)\sigma(-v^T z) \} \\ 
 &= \argmin_{\Theta} - \log \frac{1}{1+\exp{(-u^Tz)}} - \log \frac{1}{1+\exp{(v^Tz)}} \\
 & = \argmin_{\Theta} \log (1+\exp{(-u^Tz)}) + \log (1+\exp{(v^Tz)}) \\
 & \approx \argmin_{\Theta} \exp{(-u^Tz)} + \exp{(v^Tz)} \\
 & \approx \argmin_{\Theta}\{v^Tz- u^Tz + 2, 0\}\\
 & = \argmin_{\Theta} \{(\alpha v^Tz- \alpha u^Tz + 2\alpha), 0\}\\
\end{aligned}
\end{equation}
where $\alpha$ is a positive scalar, $\Theta$ represents learnable parameters of $u, v, $ and $z=\Phi(x)$. The above two approximations are achieved by $u^Tz \to + \infty$ and $v^Tz \to - \infty$.

As for the $\mathcal{L}_{\text{CSCE}}$, the (ratio form) contrastive term $\log \frac{\exp{(u^T z)}}{\exp{(u^T z)} + \exp{(v^T z)}}$ can be approximated as:
\begin{equation}
\begin{aligned}
 \argmin_{\Theta} & - \log \frac{\exp{(u^T z)}}{\exp{(u^T z)} + \exp{(v^T z)}} \\
 & = \argmin_{\Theta} \log (1+\exp{((v-u)^T z)})\\
 & \approx \argmin_{\Theta} \exp{((v-u)^T z)} \\
& \approx \argmin_{\Theta} \{v^T z- u^T z + 1, 0\}\\
& = \argmin_{\Theta} \{(\alpha v^T z- \alpha u^T z + \alpha), 0\}\\
\end{aligned}
\end{equation}
where the approximations are  achieved by $(v-u)^T z \to - \infty$ and $\alpha$ is a positive scalar.

Though different forms, both contrastive cross entropy losses  $\mathcal{L}_{\text{CBCE}}$ and $\mathcal{L}_{\text{CSCE}}$  try to make the distance between $z$ and the targeted anchor $u$ smaller than the distance between $z$ and negative anchor $v$. Similar argument applies to the $\mathcal{L}_{\text{SCR}}$ as the ratio form contrastive term  $\mathcal{L}_{\text{CSCE}}$. This is the major reason why all the above losses are named as \mytag{contrastive}.

\subsection{Generalization to Multi-label Classification}
\label{sec:model:multi-label}
We further generalize the above binary classification losses to  multi-label classification losses. A typical clinical prediction application is phenotyping which tries to predict the existences of multiple clinical conditions. We model multi-label classification as solving multiple binary classifications simultaneously. Here we define our general multi-label form of $\mathcal{L}_{\text{Supervised Contrastive} }$ as follows:
\begin{equation}
\label{eq:multi-main}
\begin{aligned}
& \mathcal{L}_{\text{Supervised Contrastive} }^{c} = \\ & \frac{1}{C}\sum_{c=1}^{C} \mathcal{L}_{\text{Contrastive Cross Entropy} }^{c}+ 
\lambda \mathcal{L}_{\text{Supervised Contrastive Regularizer}}^{c}
\end{aligned}
\end{equation}
where $C$ is the number of classes. Equation~\ref{eq:scl-main} is a special case of Equation~\ref{eq:multi-main} when $C=1$.

Based on the aforementioned contrastive cross entropy losses $\mathcal{L}_{CBCE}$, $\mathcal{L}_{CSCE}$ (sec.~\ref{sec:model:cce}), and the supervised contrastive regularizer $\mathcal{L}_{SCR}$ (sec.~\ref{sec:model:scr}), here we propose following two versions of our general supervised contrastive loss:
\bit
\item
Our general multi-label form $\mathcal{L}_{\text{CBCE}} + \lambda \mathcal{L}_{\text{SCR}}$
 is:
\begin{equation}
\small
\label{eq:multi-cbce}
\begin{aligned}
& \frac{1}{C}\sum_{c=1}^{C}\mathcal{L}_{CBCE}^{c} + \lambda \mathcal{L}_{SCR}^{c} \\
& = \frac{-1}{CN} \sum_{c=1}^{C} \sum_{i=1}^{N} \Big\{  y_{i,c} \log \sigma(u_c^T z_i)\sigma(-v_c^T z_i) + (1-y_{i,c}) \log \sigma(v_c^T z_i)\sigma(-u_c^T z_i) \\
&+ \frac{\lambda}{N_{y_{i,c}}-1} \sum_{j=1}^{N} \mathbf{1}_{i \neq j} \mathbf{1}_{y_{i,c}=y_{j,c}} \log \frac{\exp (\text{sim}(z_i, z_j) / \tau)}{\sum_{k=1}^{N} \mathbf{1}_{i \neq k} \exp (\text{sim}(z_i, z_k) / \tau)} \Big\}
\end{aligned}
\end{equation}

\item
Our general multi-label form $\mathcal{L}_{\text{CSCE}} + \lambda \mathcal{L}_{\text{SCR}}$
is:
\begin{equation}
\small
\label{eq:multi-csce}
\begin{aligned}
& \frac{1}{C}\sum_{c=1}^{C}\mathcal{L}_{CSCE}^{c} + \lambda \mathcal{L}_{SCR}^{c} \\
& = \frac{-1}{CN} \sum_{c=1}^{C} \sum_{i=1}^{N} \Big\{ y_{i,c} \log \frac{\exp{(u_c^T z_i)}}{\exp{(u_c^T z_i)} + \exp{(v_c^T z_i)}}  \\
&\quad \quad  \quad \quad  + (1-y_{i,c}) \log \frac{\exp{( v_c^T z_i )}}{\exp{(u_c^T z_i)} + \exp{(v_c^T z_i)}} \\
& + \frac{\lambda}{N_{y_{i,c}}-1} \sum_{j=1}^{N} \mathbf{1}_{i \neq j} \mathbf{1}_{y_{i,c}=y_{j,c}} \log \frac{\exp (\text{sim}(z_i, z_j) / \tau)}{\sum_{k=1}^{N} \mathbf{1}_{i \neq k} \exp (\text{sim}(z_i, z_k) / \tau)} \Big\}
\end{aligned}
\end{equation}
\eit
It is worthwhile to mention that the above two multi-label classification losses encompass binary-classification losses as special cases when $C=1$. For simplicity, we use general form $\mathcal{L}_{\text{Supervised Contrastive} }  = \mathcal{L}_{\text{Contrastive Cross Entropy} }  + \lambda \mathcal{L}_{\text{Supervised Contrastive Regularizer}} $ to denote both binary and multi-label cases.

\subsection{Summary} 
\label{sec:model:all}
\begin{algorithm}[!th]
\SetAlgoLined
 \textbf{Input:} Data $X=\{x_i\}_{i=1}^{N}$, labels $Y = \{y_i\}_{i=1}^{N}$ \\
 \textbf{Output:} $\Phi_{\Theta}$:targeted neural encoder for $X$, $U=\{u_c\}_{i=1}^{C}, V=\{v_c\}_{i=1}^{C}$: positive and negative anchors for each of $C$ classes\\
\For{ each epoch }
{
    \mytag{Step 1:} Sampling mini-batch $X=\{x_i\}_{i=1}^{n}$ \\
    \mytag{Step 2:} Generating data representations $Z=\{z_i\}_{i=1}^{n} = \Phi(\{x_i\}_{i=1}^{n})$ \\
    \mytag{Step 3:} Computing  the supervised contrastive loss $\mathcal{L}_{\text{Supervised Contrastive} }  = \mathcal{L}_{\text{Contrastive Cross Entropy} }  + \lambda \mathcal{L}_{\text{Supervised Contrastive Regularizer}} $  by Eq.~\ref{eq:multi-cbce} or Eq.~\ref{eq:multi-csce} \\
    \mytag{Step 4:} Updating $\Theta, U, V$ by minimizing above loss. 
}
\quad \textbf{Return:} $\Phi_{\Theta}$, $U, V$\\
 \caption{ The learning framework of our \model \label{alg:scehr}}
\end{algorithm}

We summarize the overall learning framework of  our \model  in Algorithm~\ref{alg:scehr}. We illustrate the main idea of our \model in Figure~\ref{fig:model}. The major outputs of algorithms are the targeted neural encoder $\Phi_{\Theta}$ for $X$, the learned positive anchors $U=\{u_c\}_{i=1}^{C}$   for each of $C$ classes, the learned negatives anchors $V=\{v_c\}_{i=1}^{C}$  for  each of $C$ classes $V=\{v_c\}_{i=1}^{C}$.
The  predicted probability of data $i$ belonging to the positive cases of class $c$ (e.g. the predicted risk of in-hospital mortality for mortality prediction task and $c=1$ represents positive/mortality) are $\sigma(u_c^T z_i) / (\sigma(u_c^T z_i)  + \sigma(v_c^T z_i))$  and $ \exp{(u_c^T z_i)} / (\exp{(u_c^T z_i)} + \exp{(v_c^T z_i)}) $ for Eq.~\ref{eq:multi-cbce} and Eq.~\ref{eq:multi-csce} respectively. 
In general, our \model can be used for existing clinical risk prediction models which are used for binary or multi-label classifications by replacing cross entropy losses with our Eq.~\ref{eq:multi-cbce} and Eq.~\ref{eq:multi-csce}.
The PyTorch implementations of our \model are open-sourced at \code.

\section{Experiments}
\label{sec:exp}
We validate our \model on a real-world electronic health records (EHR) database, Medical Information Mart for Intensive Care (MIMI-III) \cite{johnson2016mimic}, which is publicly available.
Following  benchmarking  works \cite{harutyunyan2019multitask}, we validate our \model by answering the following questions:
\bit
\item \mytag{In-hospital mortality prediction (Sec.~\ref{sec:exp:mortality})} tries to predict in-hospital mortality states, namely a binary classification task, of ICU patients given their first 48-hour data in ICU. The early-prediction of at-risk patients is the key for patient stratification to improve healthcare results. Our question is:
Can our \model improve the performance of benchmarking models for in-hospital mortality prediction task? 

\item \mytag{Phenotyping classification (Sec.~\ref{sec:exp:phenotype})} tries to predict the existence of 25 common clinical conditions (coded by ICD-9 codes in EHR) of patients in ICU, namely a multi-label classification task, given their data in ICU with varying length of time. The phenotyping is key for diagnosis, comorbidity detection, and quality surveillance \cite{oellrich2016digital}. Our question is: Can our \model improve the performance of typical benchmarking models for phenotyping task? 

\item \mytag{Data Imbalance Analysis (Sec.~\ref{sec:exp:imbalance})}. Positive cases in the EHR data always make up a smaller proportion than the negative cases. Our question is: How will our \model perform under different levels of data imbalance?

\item \mytag{Embedding Visualization (Sec.~\ref{sec:exp:visualization})}. Our \model is supposed to pull similar data embeddings closer and push dissimilar ones apart. Our question is: What will the learned embeddings look like by our \model on the real-world EHR data?
\eit

\mytag{Datasets.} 
Following the benchmark tasks \cite{harutyunyan2019multitask} on the MIMI-III dataset \cite{johnson2016mimic}, $17$ medical concepts (including Capillary refill rate, Diastolic blood pressure, Fraction inspired oxygen, Heart Rate, etc.) observed over time are selected as features, which are further feature-engineered into $76$ dimensional medical time series data  for predictive models. As for the mortality prediction, the first $48$ hour time series are used, leading to $x_{i}  \in \mathbb{R}^{48 \times 76}$ medical time series for each patient. Besides, the latest works \cite{ma2020concare} also included additional $12$ dimensional static features based on demographics (e.g. ethnicity, gender, age, height, weight, etc.) to improve the performance. The supervised labels are $\{0, 1\}^N$ for $N$ patients. As for the phenotyping classification, the time length $T_i$ of $x_{i}  \in \mathbb{R}^{T_i \times 76}$ varies depends on the length of stay in ICU. The labels for phenotyping multi-label classification are $\{0, 1\}^{N \times 25}$.  The splitting of the train, validation, and test datasets are summarized in Table~\ref{tab:data}, and the statistics of the varying $T_i$ for phenotyping classification are summarized in Table~\ref{tab:data_length}.

\hide{
medical time series data
features 17, 64?
,labels, length

demographics for latest sota models

train:
Out[21]: DescribeResult(nobs=29250, minmax=(1, 2804), mean=86.8160341880342, variance=15343.792371703845, skewness=5.255800057690939, kurtosis=46.441685494936614)

val:
Out[23]: DescribeResult(nobs=6371, minmax=(2, 1843), mean=88.78684664887773, variance=15764.25911110662, skewness=4.458990659743004, kurtosis=27.679720180506603)

test:
Out[27]: DescribeResult(nobs=6281, minmax=(2, 1993), mean=88.75354242954944, variance=16296.498484662568, skewness=5.040061285922669, kurtosis=38.76850251265428)
}

\begin{table}[!tb] 
\centering 
\caption{Statistics of datasets. The ratio of positive cases is shown in the round brackets. The mortality data have binary labels, and the  phenotyping data have 25-dimensional multi-labels.}
\begin{tabular}
{l c c c c} 
\toprule 
 \bf{} &  \bf{\#Train} &  \bf{\#Validation} &  \bf{\#Test}\\ 
\midrule 
Mortality & 14,681 (13.53\%) & 3,222 (13.53\%) & 3,236 (11.56\%)  \\ 
Phenotyping & 29,250 (16.54\%) & 6,371 (16.31\%) & 6,281 (16.53\%)  \\ 
\bottomrule 
\end{tabular}
\label{tab:data} 
\end{table}

\begin{table}[!tb] 
\centering 
\caption{Statistics of the varying length $T_i$ of each patient in phenotyping dataset.}
\begin{tabular}
{l c c c c} 
\toprule 
 \bf{Phenotyping} &  \bf{\#Train} &  \bf{\#Validation} &  \bf{\#Test}\\ 
\midrule 
min & 1 &  2&  2 \\ 
max & 2804 & 1843 &  1993 \\
mean & 86.81 & 88.79 &  88.75 \\
std. & 123.87 & 125.56 & 127.66   \\
\bottomrule 
\end{tabular}
\label{tab:data_length} 
\end{table}


We implemented our codes by Python 3.9.1,  Pytorch-1.7.1, Cuda 10.1 and trained all the models on $1$ GeForce RTX 2080 Ti GPU and 16 CPU cores in Linux server with Ubuntu 18.04.2 LTS.  We open-source our codes at \url{https://github.com/**/SCL-EHR} and refer to \cite {johnson2016mimic} for the public MIMIC-III dataset and \cite{harutyunyan2019multitask} for the data pre-processing and benchmarking codes.


\subsection{In-hospital Mortality Prediction}
\label{sec:exp:mortality}

\hide{
BCE: 0.8536,
BCE+SCL.a0.0.bs128.wdcy0.0.epo100.TrLos0.23.VaLos0.32.ACC0.877.ROC0.8312.PRC0.4965.TstACC0.882.ROC0.8232.PRC0.4416.csv

BCE+SCL:  0.8584,
BCE+SCL.a0.009.bs1024.wdcy0.0.epo100.TrLos0.36.VaLos0.40.ACC0.885.ROC0.8443.PRC0.5251.TstACC0.896.ROC0.8510.PRC0.4791.csv

CBCE:  0.8586,
CBCE+SCL.a0.0.bs256.wdcy0.0.epo100.TrLos0.24.VaLos0.31.ACC0.882.ROC0.8432.PRC0.5236.TstACC0.892.ROC0.8437.PRC0.4792.csv

CBCE+SCL:  0.8599,
CBCE+SCL.a0.0025.bs256.wdcy0.0.epo100.TrLos0.26.VaLos0.33.ACC0.878.ROC0.8422.PRC0.5229.TstACC0.895.ROC0.8507.PRC0.4874.csv

MCE: 0.8581,
MCE+SCL.a0.0.bs256.wdcy0.0.epo100.TrLos0.24.VaLos0.31.ACC0.881.ROC0.8351.PRC0.5221.TstACC0.893.ROC0.8474.PRC0.4717.csv

MCE+SCL: 0.8598
MCE+SCL.a0.0025.bs512.wdcy0.0.epo100.TrLos0.27.VaLos0.33.ACC0.889.ROC0.8423.PRC0.5368.TstACC0.898.ROC0.8537.PRC0.4887.csv
}

\begin{table*}[!htb] 
\centering 
\caption{In-hospital mortality prediction results by benchmarking LSTM model \cite{harutyunyan2019multitask} under different losses. BCE: Binary Cross Entropy; CBCE: Contrastive Binary Cross Entropy; CSCE: Contrastive Softmax Cross Entropy; SCR: Supervised Contrastive Regularizer.  We highlight the best performance w.r.t different metrics. We also report the standard deviation (std.) of bootstrapped results by re-sampling the test set 100 times with replacement in round brackets for reference.}
\begin{tabular}{l c c c c } 
\toprule 
 & \textbf{ AUROC} & \textbf{ AUPRC} & \textbf{ Accuracy} &  \textbf{ min(Se, P+)}  \\ 
\midrule 
$\mathcal{L}_{\text{BCE}}$     & $0.854 (0.010)$  & $0.483 (0.031)$ & $0.896 (0.005)$  & $0.487 (0.026)$\\ 
$\mathcal{L}_{\text{BCE}} + \lambda \mathcal{L}_{\text{SCR}}$ & $0.858 (0.009)$  & $0.489 (0.028)$  & 0.892 (0.005)  &  0.487 (0.023)   \\ 
\textbf{$\mathcal{L}_{\text{CBCE}} + \lambda \mathcal{L}_{\text{SCR}}$} & $\mathbf{0.860 (0.009)}$ & $\mathbf{0.504 (0.031)}$ & $\mathbf{0.897 (0.005)}$  &  0.482 (0.025) \\ 
\textbf{$\mathcal{L}_{\text{CSCE}} + \lambda \mathcal{L}_{\text{SCR}}$} & $\mathbf{0.860 (0.010)}$ & 0.501 (0.030) & 0.893 (0.005)  & $\mathbf{0.505 (0.024)}$  \\ 
\bottomrule 
\end{tabular}
\label{tab:mortality_lstm} 
\end{table*}

\begin{table*}[!htb] 
\centering 
\caption{In-hospital mortality prediction results by benchmarking Concare \cite{ma2020concare} model under different losses. Additional static demographic features are used in this experiment. 
}
\begin{tabular}{l c c c c } 
\toprule 
 & \textbf{ AUROC} & \textbf{ AUPRC} & \textbf{ Accuracy} &  \textbf{ min(Se, P+)}  \\ 
\midrule 
$\mathcal{L}_{\text{BCE}}$    & $ 0.864 (0.010)$  & $ 0.500 (0.027)$ & $ 0.899 (0.005)$  & $ 0.484 (0.022)$\\ 
$\mathcal{L}_{\text{BCE}} + \lambda \mathcal{L}_{\text{SCR}}$ & $ 0.864 (0.009)$  & $ 0.494 (0.027)$ & $ 0.901 (0.005)$  & $ \mathbf{0.500 (0.022)}$\\
\textbf{$\mathcal{L}_{\text{CBCE}} + \lambda \mathcal{L}_{\text{SCR}}$} & $\mathbf{ 0.868(0.008)}$ & $ 0.507 (0.027) $ & $\mathbf{0.903 (0.005)}$  &   $0.484 (0.021)$\\ 
\textbf{$\mathcal{L}_{\text{CSCE}} + \lambda \mathcal{L}_{\text{SCR}}$} & $\mathbf{ 0.868 (0.009) }$ & $\mathbf{0.508 (0.027)}$ &  0.902 (0.005)  & $ 0.497 (0.022) $  \\ 
\bottomrule 
\end{tabular}
\label{tab:mortality_concare} 
\end{table*}

\mytag{Setup.} The in-hospital mortality prediction, which is formulated as a binary classification problem, is always learned by optimizing binary cross entropy (BCE) loss in existing works \cite {harutyunyan2019multitask, ma2020concare}. In this task, we evaluate our \model's capability of improving benchmark models for mortality prediction by replacing the BCE loss.  

To be comparable with benchmark models, we adopt the most widely used: a) LSTM-based models (a 2-layerd LSTM model with $7,697$ learnable parameters) \cite {harutyunyan2019multitask} ;  and b) the state-of-the-art attention-based model Concare  (a complex channel-wise GRU model with attention layers and using additional static demographic features, leading to $322,706$ learnable parameters in total) \cite{ma2020concare}, and compare these models with a) their original binary cross entropy loss {$\mathcal{L}_{\text{BCE}}$} ; b) binary cross entropy loss with supervised contrastive regularizer  {$\mathcal{L}_{\text{BCE}} + \lambda \mathcal{L}_{\text{SCR}}$} ; c) our contrastive binary cross entropy loss with supervised contrastive regularizer  {$\mathcal{L}_{\text{CBCE}} + \lambda \mathcal{L}_{\text{SCR}}$} ; d) our contrastive softmax cross entropy loss with supervised contrastive reularizer  {$\mathcal{L}_{\text{CSCE}} + \lambda \mathcal{L}_{\text{SCR}}$}.  To be consistent with baseline implementations, we control for the same learning settings, including Adam optimizer \cite{kingma2014adam}  with learning rate $0.001$, dropout $0.3$,  weight decay $0$, and only  grid search for best AUROC performance among two varying hyper-parameters, namely, batch size $\{128, 256, 512, 1024\}$ and $\lambda \in [0, 0.01]$. The hidden dimensions of $Z$, namely the penultimate layer for contrastive learning regularizer are $16$ for LSTM and $32$ for Concare. We set the maximum epochs of training for LSTM and Concare are 100 and 150 respectively. We set the temperature $\tau=0.1$ for all the following experiments.

We evaluate the performance of this binary classification by the widely-adopted benchmark metrics, including \textit{AUROC} which is the area under the receiver operating characteristic curve; \textit{AUPRC} which is the area under the precision and recall (also known as sensitivity) curve; \textit{Accuracy} which is the ratio of correctly predicted cases to the total cases; and \textit{min(Se, P+)} which is the upper bound of the minimum of different sensitivity and precision pairs.

\mytag{Results.} Table~\ref{tab:mortality_lstm} and Table~\ref{tab:mortality_concare} show that our \model improves the best performance of both the benchmark LSTM model and the state-of-the-art Concare model with respect to all the four metrics for the in-hospital mortality prediction task on the MIMIC-III dataset. 
More specifically, we find both two contrastive losses \modela and \modelb outperforms $\mathcal{L}_{\text{BCE}}$ w.r.t all the metrics. The \modela achieved the best AUROC, AUPRC, Accuracy, while the \modelb achieved similar AUROC and the best min(Se, P+) for both models, regardless of the different complexity of two benchmark models. Besides, simply applying the regularizer $\lambda \mathcal{L}_{\text{SCR}}$ to $\mathcal{L}_{\text{BCE}}$ also improves the best AUROC performance of using bare $\mathcal{L}_{\text{BCE}}$ for LSTM. 

We observe similar empirical running times for different losses under the same predictive model. All the above loss functions finish 100 epochs with 256 batch size within $~3$ minutes for the LSTM-based model and $~45$ minutes for the Concare model. 

In conclusion,  \modela or \modelb improves the performance of strong benchmarking model LSTM and the state-of-the-art Concare model by replacing BCE loss. Both two supervised contrastive terms, namely $\mathcal{L}_{\text{Contrastive Cross Entropy} }$  and $\mathcal{L}_{\text{Supervised Contrastive Regularizer}} $  can introduce additional performance improvement. 

\subsection{Phenotyping Classification}
\label{sec:exp:phenotype}


\begin{table}[!htb] 
\small
\centering 
\caption{Prediction results of $25$ Phenotypes by benchmarking LSTM \cite{harutyunyan2019multitask} model under different losses. BCE: Multi-label Binary Cross Entropy; CBCE: Multi-label Contrastive Binary Cross Entropy; CSCE: Multi-label Contrastive Softmax Cross Entropy; SCR: Multi-label Supervised Contrastive Regularizer. We highlight the best performance w.r.t different metrics. 
}
\begin{tabular}{l c c c } 
\toprule 
 & \textbf{\makecell{Micro \\ AUROC}} & \textbf{\makecell{Macro \\ AUROC}} & \textbf{\makecell{Weighted \\ AUROC}} \\ 
\midrule 
$\mathcal{L}_{\text{BCE}}$    & $ 0.822$  & $0.772 $ & $0.758 $  \\ 
$\mathcal{L}_{\text{BCE}} + \lambda \mathcal{L}_{\text{SCR}}$ & $ \mathbf{0.824} $  & $\mathbf{0.775}$ & $\mathbf{0.761}$  \\
\textbf{$\mathcal{L}_{\text{CBCE}} + \lambda \mathcal{L}_{\text{SCR}}$} & $0.823$ & $0.774 $ & $\mathbf{0.761 }$  \\ 
\textbf{$\mathcal{L}_{\text{CSCE}} + \lambda \mathcal{L}_{\text{SCR}}$} & $\mathbf{ 0.824  }$ & $0.774$ &  $\mathbf{0.761}$  \\ 
\bottomrule 
\end{tabular}
\label{tab:phenotype} 
\end{table}

\hide{
BCE: pytorch_states/BCE/LSTM.i76.h256.L1.c25.D0.3.BCE+SCL.a0.0.bs256.wdcy0.0.epo30.Val-AucMac0.7690.AucMic0.8202.Tst-AucMac0.7717.AucMic0.8215.pt

BCE + SCR: pytorch_states/BCE/LSTM.i76.h256.L1.c25.D0.3.BCE+SCL.a0.01.bs256.wdcy0.0.epo34.Val-AucMac0.7716.AucMic0.8216.Tst-AucMac0.7748.AucMic0.8235.pt

CBCE + SCR: ./pytorch_states/CBCE/LSTM.i76.h256.L1.c50.D0.3.CBCE+SCL.a0.003.bs256.wdcy0.0.epo31.Val-AucMac0.7716.AucMic0.8214.Tst-AucMac0.7744.AucMic0.8234.pt

MCE + SCR: ./pytorch_states/MCE/LSTM.i76.h256.L1.c50.D0.3.MCE+SCL.a0.01.bs512.wdcy0.0.epo44.Val-AucMac0.7720.AucMic0.8219.Tst-AucMac0.7740.AucMic0.8236.pt
}

\mytag{Setup.} The phenotyping, which is formulated as a multi-label classification problem, is learned by optimizing the mean of multiple binary cross entropy losses (BCE) in existing benchmarking models \cite{harutyunyan2019multitask}. In this task, we evaluate our \model's ability to improve the benchmarking phenotyping models by replacing the BCE loss. 

We examined the LSTM-based model (a 1-layerd LSTM model with $348,441$ learnable parameters) \footnote{We choose standard LSTM benchmarking model because different LSTM benchmarks in \cite {harutyunyan2019multitask} have similar auroc performance, and the state-of-the-art Concare \cite{ma2020concare} can not be applied to time series with varying length.} \cite {harutyunyan2019multitask} under different losses, including a)  \textit{multi-label}   cross entropy loss {$\mathcal{L}_{\text{BCE}}$} ; b) \textit{multi-label}   cross entropy loss with \textit{multi-label} supervised contrastive  regularizer  {$\mathcal{L}_{\text{BCE}} + \lambda \mathcal{L}_{\text{SCR}}$} ; c) our \textit{multi-label} contrastive binary cross entropy loss with \textit{multi-label} supervised contrastive regularizer  {$\mathcal{L}_{\text{CBCE}} + \lambda \mathcal{L}_{\text{SCR}}$} ; d) our \textit{multi-label}  contrastive softmax cross entropy loss with \textit{multi-label}  supervised contrastive reularizer  {$\mathcal{L}_{\text{CSCE}} + \lambda \mathcal{L}_{\text{SCR}}$}.  
We evaluate multi-label classification performance by standard metrics including \textit{Micro-AUROC}, \textit{Macro-AUROC}, and \textit{weighted-AUROC} \cite{aurocurl}. We adopt the same setting  for consistency, including Adam optimizer with learning rate $0.001$, dropout $0.3$,  weight decay $0$, and we  grid search for best micro-AUROC performance among two varying hyper-parameters, namely, batch size $\{128, 256, 512, 1024\}$ and $\lambda \in [0, 0.01]$. The hidden dimension of $Z$, namely the penultimate layer for contrastive learning regularizer is $256$.

\mytag{Results.} Table~\ref{tab:phenotype} reports different AUROC scores, we find that our \model improves benchmarking LSTM models w.r.t all the metrics. More specifically, our \modelb and applying $\mathcal{L}_{\text{SCR}}$ directly to BCE loss achieved the best performance, indicating the benefits of introducing supervised contrastive terms.

\subsection{Data Imbalance Analysis}
\label{sec:exp:imbalance}
\begin{figure}[!ht]
\centering
\includegraphics[width=0.5\textwidth]{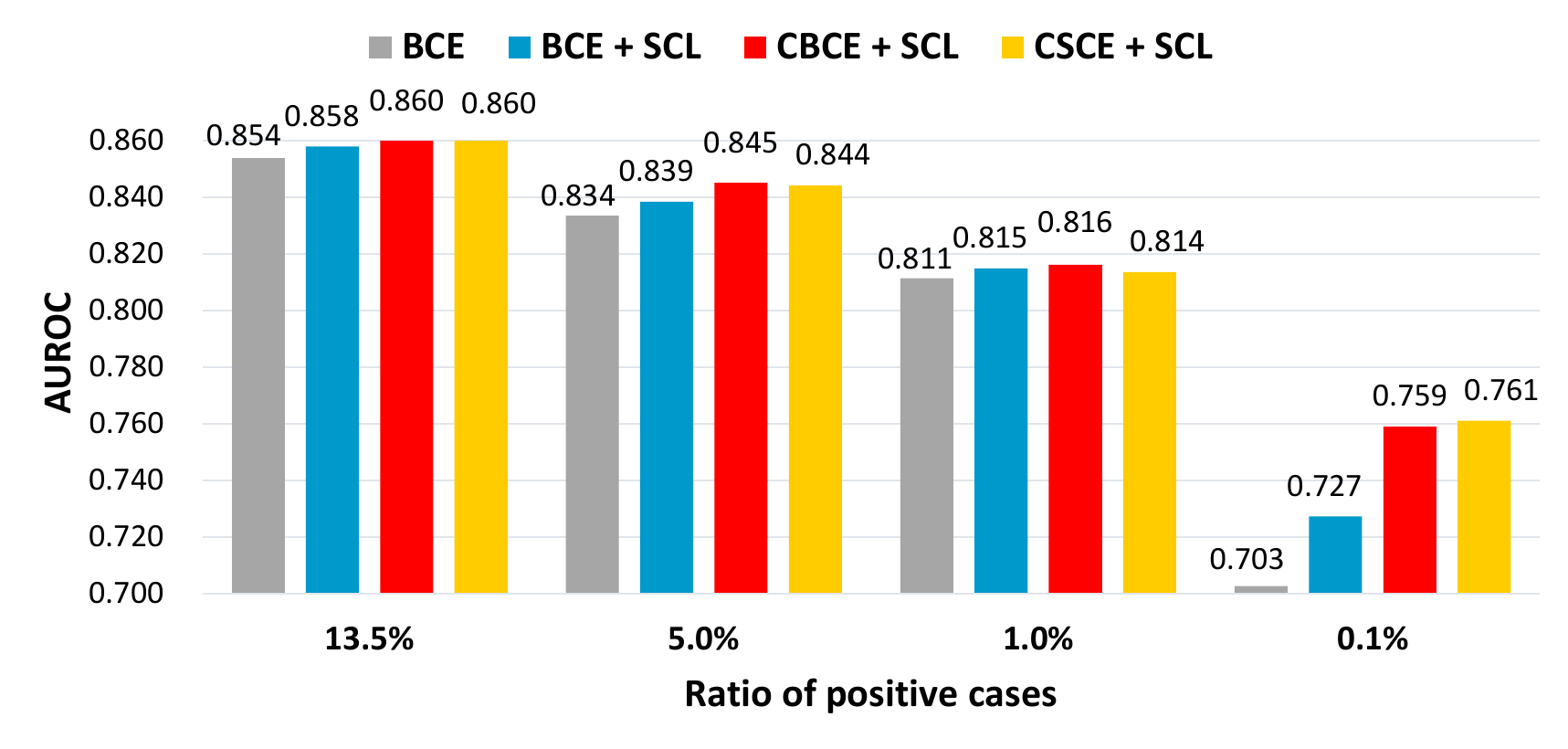}
\caption{ 
In-hospital mortality prediction under different data imbalance levels. 
\label{fig:imbalance}}
\end{figure}

\mytag{Setup.} 
We further investigate the performance of our loss functions when the number of positive cases in the training data is imbalanced at different levels. We studied the in-hospital mortality prediction by the benchmarking LSTM model. As shown in Table~\ref{tab:data}, the original ratio of positive cases in the training dataset is $13.53\%$. We downsample the training data with different levels of positive cases, namely, $5\%$, $1\%$, and $0.1\%$, and keep the test data the same. The number (with the ratio of positive cases in the round brackets) of patients in down-sampled training datasets are $13,374 \ (5\%)$ , $12,825 \  (1\%)$, $12,708 \ (0.1\%)$, respectively.
Follow the same experimental setting as section~\ref{sec:exp:mortality},  we search the best AUROC performance on the hyper-parameter space spanned by batch size $\{128, 256, 512, 1024\}$ and $\lambda \in [0, 0.01]$.

\mytag{Results.}
We report the AUROC achieved by different losses under different data imbalance levels (the ratio of positive cases) in Figure~\ref{fig:imbalance}.
We find consistent improvements of  our \modela and \modelb over the BCE loss under different imbalance levels. Besides, introducing the self-supervised regularizer to BCE also improves, but not as significant as \modela and \modelb. When the prevalence of positive cases is very rare, say $0.1 \%$, we find that our \modela and \modelb outperforms BCE a lot.

In conclusion, our experimental result implies that when the focused clinical outcome is rare (e.g. rare diseases) in EHR datasets, namely, a very small fraction of positive cases among the total population,  replacing the BCE loss by our \modela and \modelb can improve binary classification performance.

\subsection{Embedding Visualization}
\label{sec:exp:visualization}
\begin{figure}[!tb]
\centering
\subfigure[ BCE ]{
\includegraphics[width=0.225\textwidth]{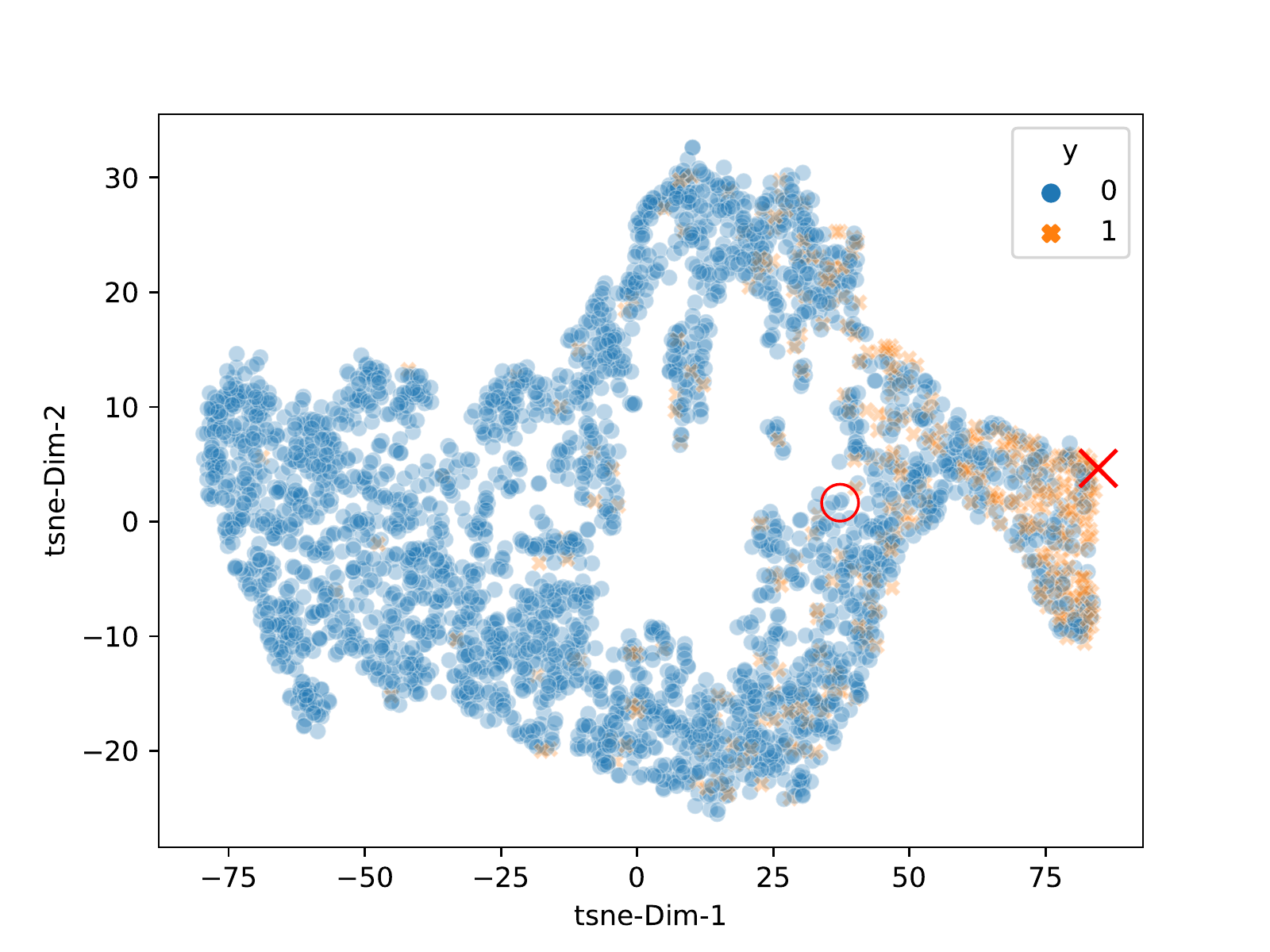}
\label{fig:syn_exp1}
}
\subfigure[BCE + SCL]{
\includegraphics[width=0.225\textwidth]{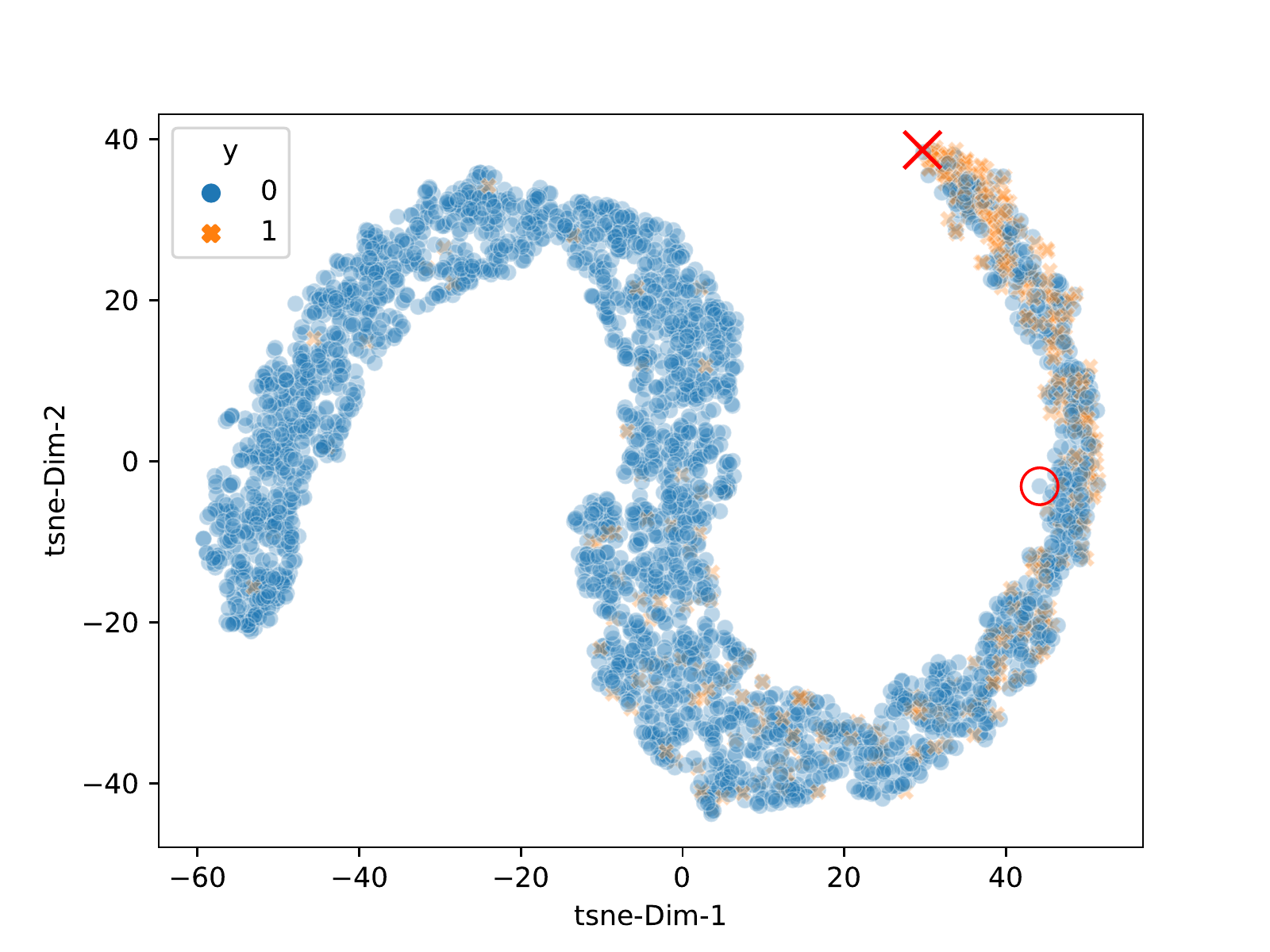}
\label{fig:syn_exp2}
}
\subfigure[ CBCE + SCL]{
\includegraphics[width=0.225\textwidth]{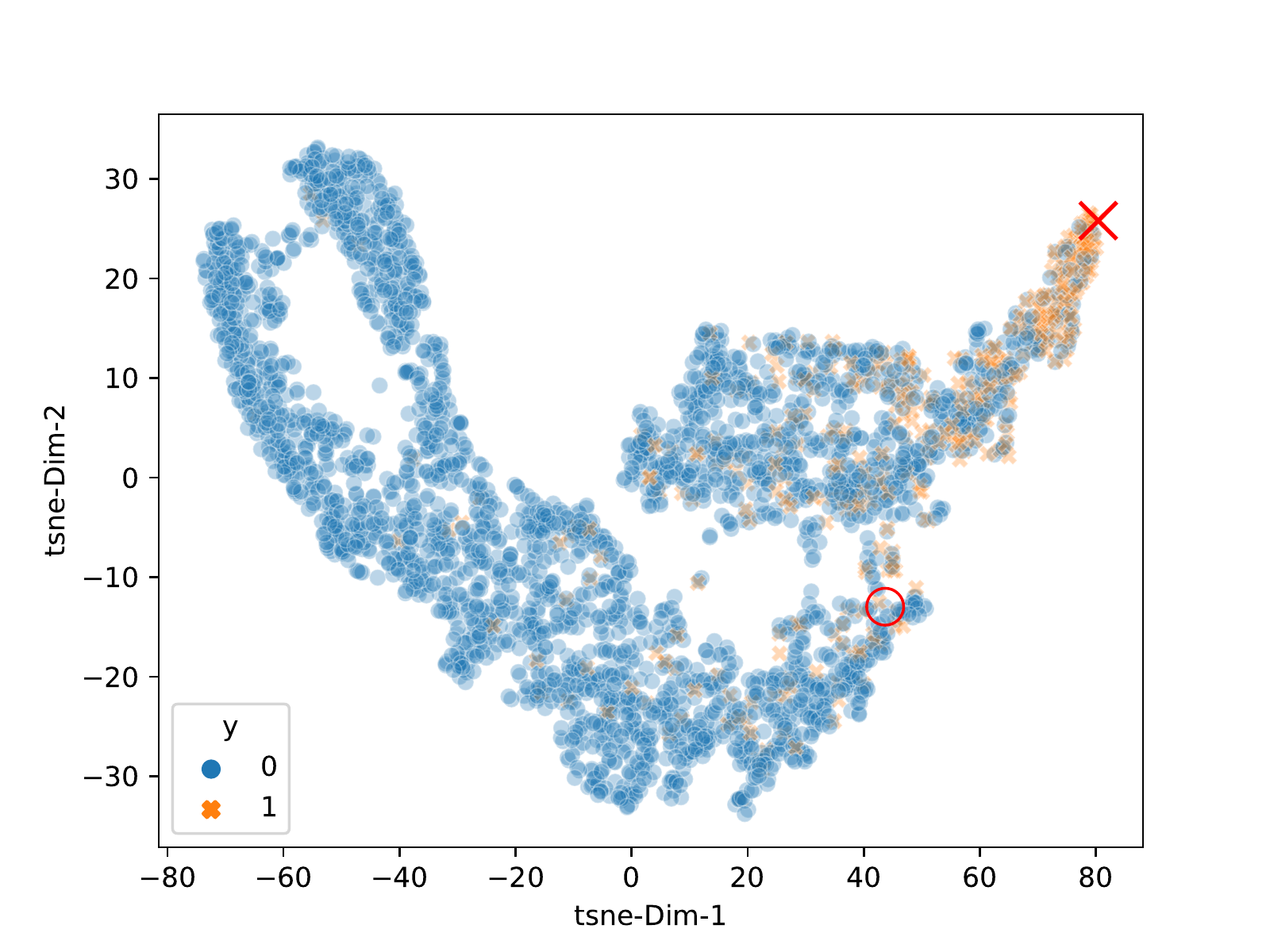}
\label{fig:syn_exp3}
}
\subfigure[CSCE + SCL]{
\includegraphics[width=0.225\textwidth]{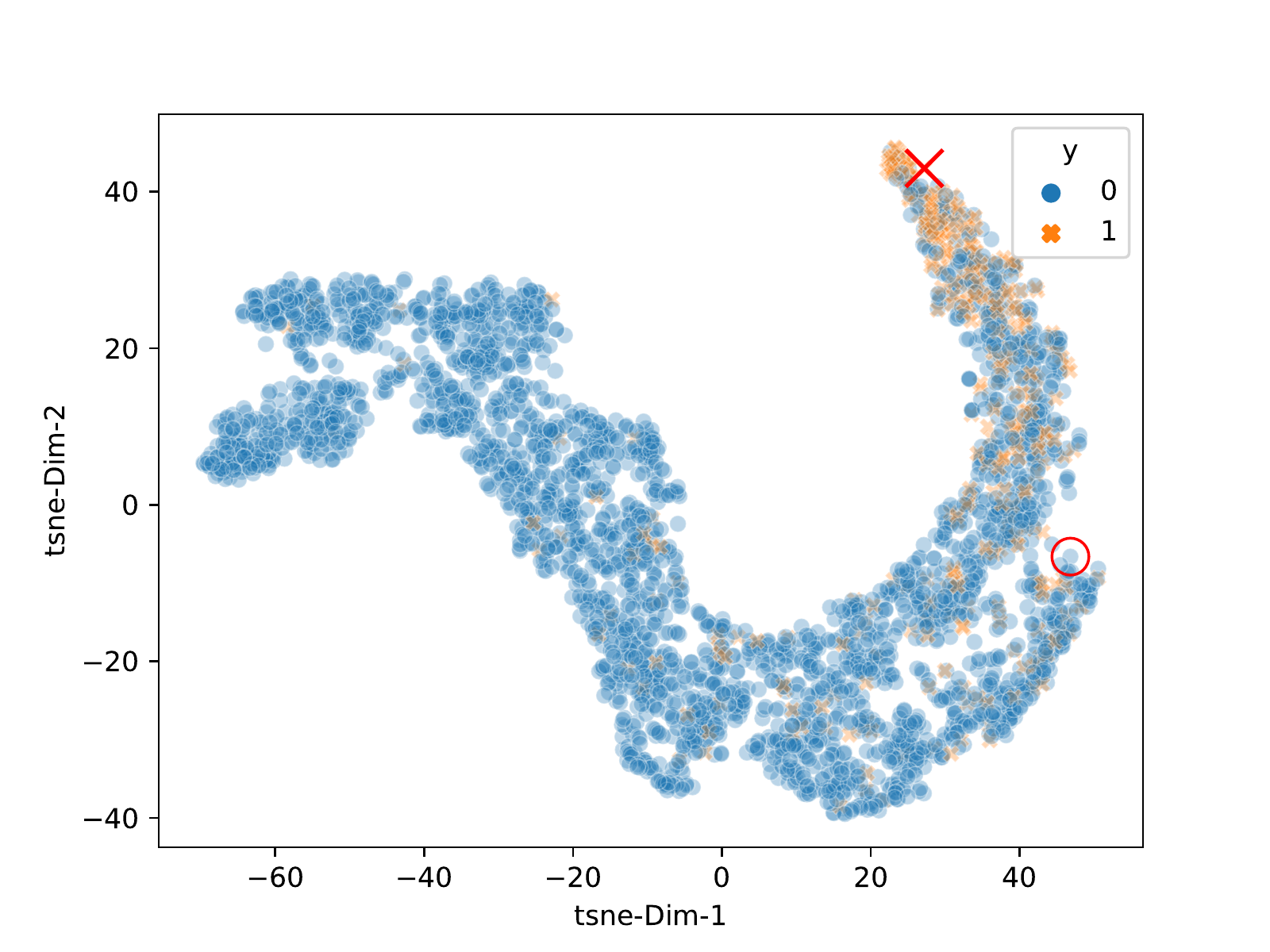}
\label{fig:syn_exp4}
}
\caption{ 
t-SNE plots of patient's embedding representations learned by the same LSTM-based mortality predictive model under BCE and different supervised contrastive losses on the test dataset. Orange crosses and  blue dots represent the positive and negative cases respectively. The positive cases account for 11.56\% of the total population. We highlight the learned positive anchor by a red cross and the negative anchor by a red dot.
\label{fig:tsne}}
\end{figure}

\mytag{Setup.}
We here try to visualize embedding representations of each patient in the test dataset learned by different losses to illustrate the effect of supervised contrastive terms.  All the representations are learned by the same  LSTM-based mortality predictive model as discussed in Section~\ref{sec:exp:mortality} under different losses, including a) the BCE loss {$\mathcal{L}_{\text{BCE}}$} ; b) BCE loss with supervised contrastive regularizer  {$\mathcal{L}_{\text{BCE}} + \lambda \mathcal{L}_{\text{SCR}}$} ; c) contrastive binary cross entropy loss with supervised contrastive regularizer  {$\mathcal{L}_{\text{CBCE}} + \lambda \mathcal{L}_{\text{SCR}}$} ; d) contrastive softmax cross entropy loss with supervised contrastive reularizer  {$\mathcal{L}_{\text{CSCE}} + \lambda \mathcal{L}_{\text{SCR}}$}.  We control for batch size 256 for all the learning processes. We plot the 16-dimensional hidden representations $Z$ by t-SNE \cite{van2008visualizing}  with 50 perplexity under 1000 iterations. The t-SNE is initialized by PCA as suggested in \cite{Kobak_Linderman_2021}.

\mytag{Results.} We show embedding visualizations in Figure~\ref{fig:tsne}. Compared with the BCE plot (Figure~\ref{fig:tsne}a), we find that all the loss functions with supervised contrastive terms (Figure~\ref{fig:tsne}b-d) better squeeze positive samples near the red cross and negative samples near the red circle, implying their ability to pull representations with the same label closer and push representations with different labels apart. What's more, compared with {$\mathcal{L}_{\text{BCE}} + \lambda \mathcal{L}_{\text{SCR}}$},  our \modela and \modelb show more complex structures and at the same time a relatively good gap between classes, which are possible reasons accounting for their better performance. Visual inspection implies best class separation by our \modela in Figure~\ref{fig:tsne}c among others, which is consistent with the best AUROC achieved by \modela. Besides, we can also find many points that are located among data clusters with different labels, indicating the intrinsic difficulty in clinical risk predictions with longitudinal EHR data \cite{bellamy2020evaluating}.

\section{Conclusion}
\label{sec:conclusion}

In this paper, we propose a general supervised contrastive loss form $\mathcal{L}_{\text{Contrastive Cross Entropy} }  + \lambda \mathcal{L}_{\text{Supervised Contrastive Regularizer}}$ for solving both binary classification
and multi-label classification in a unified framework for clinical risk prediction using EHR data. 
Our proposed loss improves the performance of strong  baselines and even state-of-the-art models on benchmarking clinical risk prediction using real-world longitudinal EHR data, works well with extremely imbalanced data, and can be easily used to existing clinical risk predictive models by replacing their (binary or multi-label) cross entropy loss. Our Pytorch code is released at \code . For future work,  more instances of the above supervised contrastive loss can be proposed. More clinical risk predictive models, EHR datasets, and self-supervised data augmentation techniques for longitudinal EHR data need further investigation.









\section*{Acknowledgement}
This work was supported by NSF 1750326, ONR N00014-18-1-2585 and NIH RF1AG072449. The authors would also like to acknowledge the support from Google Faculty Research Award and Amazon Web Services Machine Learning for Research Award.



{

\bibliographystyle{IEEEtran}
\bibliography{reference}

}

\end{document}